\pdfoutput=1
\documentclass[11pt]{article}

\usepackage[]{ACL2023}

\usepackage{times}
\usepackage{latexsym}
\usepackage{graphicx}
\usepackage[T1]{fontenc}
\usepackage[utf8]{inputenc}

\usepackage{microtype}

\usepackage{inconsolata}
\usepackage{booktabs}

\title{Zero Resource Cross-Lingual Part Of Speech Tagging}

\author{Sahil Chopra \\ Saarland University \\
  \texttt{sach00002@uni-saarland.de} }

\begin{document}
\maketitle
\begin{abstract}

Part of speech tagging in zero-resource settings can be an effective
approach for low-resource languages when no
labeled training data is available. Existing systems use two main techniques for POS tagging i.e. pretrained multilingual large language models(LLM) or project the source language labels into the zero resource target language and train a sequence labeling model on it. We explore the latter approach using the off-the-shelf alignment module and train a hidden Markov model(HMM) to predict the POS tags. We evaluate transfer learning setup with English as a
source language and French, German, and Spanish as target languages for
part-of-speech tagging. Our conclusion is that projected alignment data in zero-resource language can be beneficial to predict POS tags.

\end{abstract}

\section{Introduction}

Over the last few years, supervised machine learning methods have set higher benchmarks for numerous NLP tasks \cite{wang-etal-2018-glue}. However, their success relies heavily on annotated data specific to the field, which is not always readily available. As a result, for various application domains and less-resourced languages, other Machine Learning techniques must be created to handle un-annotated or partially annotated data.
Currently, the most effective method for a significant portion of natural language processing tasks is to fine-tune pre-existing models using labeled data that is tailored to the specific task. Regrettably, this task-specific labeled data is often unavailable, particularly for low-resource language.  One of the possible solution is fine-tuning a cross-lingual multilingual pre-trained language models \cite{conneau-etal-2020-unsupervised,Devlin2019BERTPO}, using available data from some source language to model the phenomenon in a different target language for which
labeled data does not exist. However, the similarity between source
and target language impacts performance, therefore cross-lingual transfer should not be evaluated using only a single presupposed source language, especially if training sets in multiple languages are available. In order, to limit the scope of this project, we restrict are experiments to a single source language to establish the validity of our hypothesis.

 To address the issue of insufficient annotated data, another commonly used approach is to utilize parallel data. This involves pairing a text in a language with abundant resources with its equivalent text in a less-resourced language. By transferring labels from the resource-rich language to the less-resourced language, it is possible to acquire imperfect, but still valuable, annotations that can be used to train a model for the less-resourced language in a distant supervised manner \cite{ganchev-etal-2009-dependency,yarowsky-etal-2001-inducing}.

The aim of this project is to investigate HMM performance on a part of speech tagging trained using an artificially generated corpus of language B using language A where A is a labeled resource-rich language and B is a low-resource or unannotated language.

  The rest of this paper is organised as follows: Sections 2 explains the three main components of this project, which
are constructing translation, the projection labels by implementing word alignments and training HMM. Section 3 describes in detail the data used. Section 4 reports the HMM performance trained on the generated data(GD) and annotated data(AD). Section 5 presents a discussion of this project.

\section{Methodology}

\begin{figure*}[h!]
    \centering
    \includegraphics[scale = 0.8]{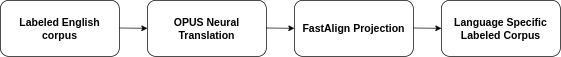}
   \caption{The pipeline for creating a data from English}
    \label{fig:corpus}
\end{figure*}

This section presents the experimental setup and functions in the script utilized to obtain the various models in our experiment. To create the corpus in the target language, and perform POS Tagging, we first have to translate and transfer the labels of the source corpus to the target as shown in the Figure \ref{fig:corpus}. We follow the same procedure as in \cite{garcia-ferrero-etal-2022-model}. The implementation of \citet{garcia-ferrero-etal-2022-model} only works for Name Entity Recognition. We had to modify the projection and alignment model to fit our use case.

First, the corpus is translated with the selected machine translation systems and then annotated tags are projected from the original to translated data.

\subsection*{OPUS Translation}

The first step in the generation of the corpus in French, German and Spanish is to translate the data in hand. In order to do so, the main objective was to find a good-quality machine translation
(MT) system. The corpus was translated by freely available OPUS-MT \cite{tiedemann-thottingal-2020-opus} MT systems. The main problems with other MT systems were that some
expressions were not translated at all and it was comparably slow or came with resource quota.  In total, 12543 sentence pairs from the English train set were translated.

\subsection*{Corpus Alignment}

Word alignment is a method in machine translation widely utilized for annotation projection. It is the natural language processing task of localising translation association among the words in a bitext, resulting in a many to many graph structure between the two sides of the bitext, with an link between two words if and only if they are translations of one another. It is used as a step to transfer labels of gold-annotated data to its translation. There exist numerous methods for word alignments. With the advent of deep learning more precise word alignment techniques have been
developed, such as aligners for cross-lingual sequence tagging  fastAlign \cite{Dyer2013ASF} and SimAlign \cite{jalili-sabet-etal-2020-simalign}.  For SimAlign we ran the model with 1.0 null align rate , no distortion rate and the itermax as a matching method. In our analysis , fastAlign creates better alignment with least number of NULL assignment. We drop the words that are not aligned to any POS tags to reduce the noisy data while learning the hidden representation.

\subsection*{Easy Projection}

Taking inspiration from \cite{garcia-ferrero-etal-2022-model}, whenever a word from the source sentence is aligned with a word from the target sentence, the target word is assigned the same category as the corresponding word in the source sentence. There are two special cases handled in projection, split annotations and annotation collision. In the first scenario, when a word in the source sentence lacks alignment, it may cause the labeled sequence to be split into multiple sequences in the target sentence. If the gap between these sequences is only one word, they are merged together.  In the annotation collision case, a word in the target sentence is aligned to two different labelled sequences in the source language. We diverge our approach in second scenario where we take first the first occurrence of projected target label if they are of different category instead of just consider the one with the longest length. This helped in retaining aligned next sub sequence.

\subsection*{Hidden Markov Model}

An HMM is a probabilistic sequence model with a  given a sequence of units (words,letters, morphemes, sentences, whatever), it computes a probability distribution over possible sequences of labels and chooses the required label sequence.  The use of a Markov chain involves a significant assumption, which is that only the current state is relevant for predicting future outcomes in the sequence. Previous states have no effect on the future, except through their influence on the current state.

 Consider a sequence of state variables $q_{1}, q_{2}, \ldots, q_{i}$. A Markov model entails the assumption on Markov probabilities of this sequence that when we are predicting the next sequence , the already occurred sequence doesn't matter, only the present.

Markov Assumption: $P\left(q_{i}=a \mid q_{1} \ldots q_{i-1}\right)=P\left(q_{i}=a \mid q_{i-1}\right)$

 A Markov chain is specified by the following components:\\
$\begin{array}{ll}Q=q_{1} q_{2} \ldots q_{N} \end{array}$ \\
$ A=a_{11} a_{12} \ldots a_{N 1} \ldots a_{N N}$ \\
$O=o_{1} o_{2} \ldots o_{T}$\\
$B=b_{i}\left(o_{t}\right)$\\
$ \pi = \pi_{i} \ldots \pi_{N}$

where $Q$ is a set of $N$ states, $A$ is a transition probability matrix , each  $ a_{i j}$ representing the probability of moving from state i to state j, $O$ a sequence of observations, $B$ is a sequence of observation likelihoods, also called emission probabilities, each expressing the probability of an observation $o_{t}$ being generated
from a state $q_{t}$ and $\pi$ an initial probability distribution over states. 
\subsection*{Viterbi}
We use the Viterbi algorithm for decoding HMMs as given in the Figure \ref{fig:viterbi}.  Given an observation sequence and
an HMM , the algorithm returns the state path through the HMM that assigns maximum likelihood to the observation sequence. The intial parameter for viterbi algorithm is calculated as follows:

initial probabilities: the probability that a sentence starts with tag $q_{j}$ \\
transition probabilities: probability of a tag $q_{j}$ given that the previous tag in this sequence was $q_{i}$

emission probabilities: probability of a word $O$ given the tag $q_{j}$.

Our tagger initially failed to produce output for sentences that contain words it haven’t seen during training. In order to implement
better unknown word handling, we use a smoothing technique. Whenever there is no emission data for the word, we replace $B$ with the $1/f(q)$ where $f()$ returns the frequency of the state.

\section{Data}

We conducted experiments using Universal Dependencies 2.8 \cite{11234/1-3687} dataset. The manually annotated data contains for 114 languages; among these all have test data and 75 languages have training data. Our training datasets consist of at least 125 samples. As a result, there 105 languages that can used
as a target languages, of which 65 can also serve as source languages since they have training data.

\begin{table}[!h]
\centering
  \begin{tabular}{lccccc}
     \toprule
     \# Language & Train & Test   
     \\ \midrule
En(ud\_english-ewt)  & 
12543 & -
\\
Fr(Fr\_gsd-ud)   & 16341 & 1000
 \\
De(De\_gsd-ud) & 15590 & 1000 
\\
Es(Es\_gsd-ud) & 16013 & 1000 
\\
\bottomrule
   \end{tabular}%
\caption{Number of samples per language }
\label{tab:DataDist}
\end{table}

We use English UD\_English-EWT dataset to generate cross lingual data in other languages and pud-ud-test dataset in French, German and Spanish languages to test. We choose these languages because they have large overlapping words with English \citep{dinu-etal-2015-cross}. To train HMM on gold AD, (gsd-ud) corpus in \citet{11234/1-3687} is used to compare the performance with GD.

The proposed universal limited POS tagset by \citet{petrov-etal-2012-universal} offers a practical foundation for associating different part-of-speech categories. While it is primarily an empirical approach, this tagset consists of 12 categories: NOUN (nouns), VERB (verbs), ADJ (adjectives), ADV (adverbs), PRON (pronouns), DET (determiners and articles), ADP (prepositions and postpositions), NUM (numerals), CONJ (conjunctions), PART (particles), and X (a catch-all for other categories). These labels have been selected for their common utility across languages and their applicability in various applications with multilingual needs.

\subsection*{Evaluation}

We calculate Precision, F1, and Recall scores for all categories of POS tags and average over it. Note that the files for the evaluation need to be in Conll standard format. We compare the HMM model trained over the generated corpus and over the labeled corpus in the same language. To keep the comparison fair, we restrict the number of sentences in the labeled corpus equal to the number of sentences in the generated corpus.

\section{Results}

\begin{table}[h]
\centering
\resizebox{\columnwidth}{!}{%
\begin{tabular}{lcccccc}
\toprule
\textbf{POS} & \multicolumn{3}{l}{\textbf{Generated Data}} & \multicolumn{3}{l}{\textbf{Annotated Data}} \\
               & Pr.  & Re. & \multicolumn{1}{l}{F1} & Pr.   & Re.  & \multicolumn{1}{l}{F1} \\ 
               
               \bottomrule

ADJ	&	0.67	&	0.54	&	0.59	&	0.85	&	0.79	&	0.82 \\
ADP	&	0.89	&	0.89	&	0.89	&	0.93	&	0.96	&	0.94 \\
ADV	&	0.71	&	0.47	&	0.57	&	0.96	&	0.83	&	0.89 \\
AUX	&	0.60	&	0.74	&	0.66	&	0.92	&	0.81	&	0.86 \\
CCONJ	&	0.95	&	0.99	&	0.97	&	0.96	&	1.00	&	0.98 \\
DET	&	0.88	&	0.93	&	0.91	&	0.88	&	0.98	&	0.93 \\
NOUN	&	0.77	&	0.82	&	0.80	&	0.93	&	0.88	&	0.90  \\
NUM	&	0.84	&	0.61	&	0.71	&	0.97	&	0.85	&	0.90  \\
PRON	&	0.57	&	0.67	&	0.62	&	0.70	&	0.53	&	0.60 \\
PROPN	&	0.50	&	0.39	&	0.44	&	0.63	&	0.64	&	0.63 \\
SCONJ	&	0.40	&	0.51	&	0.45	&	0.35	&	0.89	&	0.50 \\
VERB	&	0.78	&	0.68	&	0.73	&	0.89	&	0.86	&	0.88  \\
Overall & 0.71	&	0.70	&	0.70	& 0.83 &	0.83	&	0.82		

\end{tabular}%
}
\caption{Test results on Spanish language}
\label{tab:results_spanish}
\end{table}

\begin{table}[h]
\centering
\resizebox{\columnwidth}{!}{%
\begin{tabular}{lcccccc}
\toprule
\textbf{POS} & \multicolumn{3}{l}{\textbf{Generated Data}} & \multicolumn{3}{l}{\textbf{Annotated Data}} \\
               & Pr.  & Re. & \multicolumn{1}{l}{F1} & Pr.   & Re.  & \multicolumn{1}{l}{F1}        \\ \bottomrule

ADJ	&	0.73	&	0.55	&	0.63	&	0.92	&	0.80	&	0.86 \\
ADP	&	0.85	&	0.83	&	0.84	&	0.93	&	0.95	&	0.94 \\
ADV	&	0.83	&	0.49	&	0.61	&	0.94	&	0.92	&	0.93 \\
AUX	&	0.73	&	0.92	&	0.81	&	0.86	&	0.99	&	0.92 \\
CCONJ	&	0.97	&	0.98	&	0.98	&	0.99	&	1.00	&	1.00 \\
DET	&	0.89	&	0.83	&	0.86	&	0.94	&	0.98	&	0.96 \\
NOUN	&	0.75	&	0.85	&	0.80	&	0.94	&	0.95	&	0.95 \\
NUM	&	0.91	&	0.60	&	0.72	&	0.96	&	0.85	&	0.90 \\
PRON	&	0.55	&	0.77	&	0.64	&	0.83	&	0.94	&	0.88 \\
PROPN	&	0.54	&	0.40	&	0.46	&	0.86	&	0.67	&	0.75 \\
SCONJ	&	0.38	&	0.74	&	0.50	&	0.64	&	0.97	&	0.77 \\
VERB	&	0.74	&	0.66	&	0.70	&	0.94	&	0.86	&	0.90 \\
Overall & 0.74	&	0.72	&	0.71 & 0.90	&	0.91	&	0.90

\end{tabular}%
}
\caption{Test results on French language }
\label{tab:results_french}
\end{table}

\begin{table}[h]
\centering
\resizebox{\columnwidth}{!}{%
\begin{tabular}{lcccccc}
\toprule
\textbf{POS} & \multicolumn{3}{l}{\textbf{Generated Data}} & \multicolumn{3}{l}{\textbf{Annotated Data}} \\

             & Pr.  & Re. & \multicolumn{1}{l}{F1} & Pr.   & Re.  & \multicolumn{1}{l}{F1}        \\ \bottomrule

ADJ	&	0.68	&	0.56	&	0.61	&	0.78	&	0.71	&	0.74 \\
ADP	&	0.67	&	0.89	&	0.76	&	0.80	&	0.98	&	0.88 \\
ADV	&	0.83	&	0.53	&	0.65	&	0.80	&	0.72	&	0.76 \\
AUX	&	0.83	&	0.88	&	0.85	&	0.88	&	0.97	&	0.92 \\
CCONJ	&	0.99	&	0.78	&	0.87	&	0.99	&	0.74	&	0.85 \\
DET	&	0.78	&	0.86	&	0.82	&	0.78	&	0.88	&	0.83 \\
NOUN	&	0.78	&	0.77	&	0.77	&	0.91	&	0.81	&	0.86 \\
NUM	&	0.80	&	0.57	&	0.66	&	0.96	&	0.83	&	0.89 \\
PART	&	0.37	&	0.94	&	0.53	&	0.57	&	0.88	&	0.69 \\
PRON	&	0.67	&	0.72	&	0.69	&	0.64	&	0.73	&	0.68 \\
PROPN	&	0.49	&	0.39	&	0.43	&	0.52	&	0.57	&	0.54 \\
SCONJ	&	0.67	&	0.77	&	0.72	&	0.91	&	0.71	&	0.80 \\
VERB	&	0.69	&	0.63	&	0.66	&	0.88	&	0.76	&	0.81 \\
Overall	& 0.71 &	0.71	&	0.60	& 0.80 &	0.79 &	0.79

\end{tabular}%
}
\caption{Test results on German language }
\label{tab:results_german}
\end{table}

Table \ref{tab:results_spanish}, \ref{tab:results_french}, \ref{tab:results_german} reports the HMM performance in terms of Precision (Pr.), Recall (Re.) and F1 score. HMM trained for Spanish on generated data achieves a F1 score of 0.70 whereas it achieves a F1 score of 0.82 when a corpus is annotated. Similarly, HMM trained for French language on generated data achieves a F1 score of 0.71 whereas a F1 score of 0.74 when a corpus is annotated. Same is the case with German language. HMM trained on generated data achieves a F1 score of 0.71 whereas  a F1 score of 0.79 when a corpus is annotated. The HMM did not manage to outperform the one supervised with gold labels. However, Our underline assumption that the labeled data is unavailable makes the results significant. 

The gap in the performance of different training dataset widens in the case of Spanish and French language. PRON,PROPN and SCONJ have relatively low scores compared to the other tags. This observation is consistent in all the languages tested. This can be attributed to the disproportionate number of tags instances in the corpus and the corpus does not have enough instances for HMM to learn its representation.
\section{Discussion}

Our analysis demonstrates that the pre-training of both the source and target languages, along with the alignment of language families, writing systems, word order systems, and lexical-phonetic distance, have a huge influence on the cross-lingual performance. However, even with these considerations, our new approach's performance is not yet on the same level with that of a fully supervised POS tagger.
Our findings indicate that the errors occur due to incorrect or missing alignments, particularly with articles and prepositions such as "de" and "la."  Large multi-word names such as "Office national de l'immigration et de l'intégration en France”  are not tagged properly. Word aligners struggle to
correctly align articles in these complex expressions especially when a one-to-many or many-to-one alignment is required. For example,
the alignment of the word failed to correctly align “and the” with “et de”. 

Other primary errors are caused by systematic differences between the tags of test and supervised text. For example in French, , the contraction "du" is formed by combining the preposition "de" (of/from) with the determiner "le" (the). In the Universal Dependency Treebank, "du" is labeled as ADP (preposition) because it functions as a prepositional phrase marker. For example, "du pain" means "of the bread" or "from the bread". However, in Wiktionary and some other linguistic resources, "du" is categorized as a contraction of the determiner "le" (the). It represents the masculine singular form of "de + le". For example, "Je mange du pain" translates to "I'm eating some bread" or "I'm eating bread". This difference in labeling can lead to challenges when aligning "du" with its counterpart in parallel corpora or when considering its part-of-speech category. While certain differences in part-of-speech labeling can be linguistically justified, stemming from inherent disparities in language structure and usage, others appear to be the result of arbitrary annotation conventions.

\section{Conclusion}
In conclusion, part-of-speech tagging in zero-resource settings can be achieved through the use of projected alignment data, which can be an effective approach for low-resource languages where labeled training data is not available. Existing systems rely on either pretrained multilingual large language models or projecting source language labels into the zero-resource target language and training a sequence labeling model on it. This paper explores the latter approach using an off-the-shelf alignment module and training a hidden Markov model to predict POS tags. Through evaluation of transfer learning setups with English as a source language and French, German, and Spanish as target languages for POS tagging, the authors conclude that projected alignment data in zero-resource languages can be beneficial for predicting POS tags.
\bibliography{custom}
\bibliographystyle{acl_natbib}

\appendix

\section{Appendix}
\label{sec:appendix}

\begin{figure*}[h!]
    \centering
    \includegraphics[scale = 0.5]{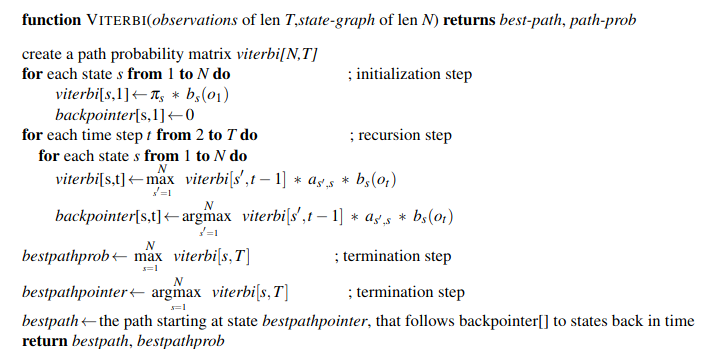}
   \caption{Viterbi Algorithm}
    \label{fig:viterbi}
\end{figure*}

\end{document}